\documentclass[11pt]{article}

\usepackage[final]{acl}

\usepackage{times}
\usepackage{latexsym}

\usepackage[T1]{fontenc}

\usepackage[utf8]{inputenc}

\usepackage{microtype}

\usepackage{inconsolata}

\usepackage{algorithm}
\usepackage{algpseudocode}
\usepackage{amsmath}
\usepackage{amsfonts}
\usepackage{amsthm}
\usepackage{amssymb}
\usepackage{booktabs}
\usepackage{multirow}
\usepackage{multicol}
\usepackage{makecell}
\usepackage{enumerate}
\usepackage{enumitem}
\usepackage{caption}
\usepackage{xcolor}
\usepackage{float}
\usepackage{graphicx}
\usepackage{xspace}
\usepackage{color}
\usepackage[table]{xcolor}

\definecolor{table_grey}{HTML}{e8e8e8}

\newcommand{\model}{EvoCtx\xspace}

\title{\textit{To Retrieve or To Think?} Cross-Boundary Context Evolution for \\ Multi-hop Complex Reasoning}

\author{
  Rubing Chen$^{1,3}$,
  ~~Jian Wang$^{2}$,
  ~~Wenjie Li$^{1}$,
  ~~Xiao-Yong Wei$^{2,1,3}$\textsuperscript{\thanks{Corresponding author.}},
  ~~Qing Li$^{1,3}$ \\
  $^{1}$ Department of Computing, The Hong Kong Polytechnic University \\
  $^{2}$ College of Computer Science, Sichuan University \\
  $^{3}$ The Hong Kong Polytechnic University Shenzhen Research Institute \\
 \texttt{rubing.chen@connect.polyu.hk ~ wangjian51@scu.edu.cn} \\
 \texttt{cswjli@comp.polyu.edu.hk ~ \{cs007.wei,qing-prof.li\}@polyu.edu.hk} 
}

\begin{document}
\maketitle
\begin{abstract}

Current context augmentation methods, such as retrieval-augmented generation, play a crucial role in bridging a model's internal knowledge boundary and external evidence for multi-hop reasoning. 
However, they often follow a rigid policy and treat external retrieval as the default action at each step. Such brute-force context expansion incurs unnecessary computational cost and may degrade reasoning performance by saturating the context with redundant or weakly relevant evidence. 
In this paper, we propose cross-boundary Context Evolution (EvoCtx), a framework that models complex reasoning as an adaptive process of boundary-aware context evolution. 
EvoCtx dynamically decides whether the next reasoning transition should cross the current evidence boundary through retrieval or refine the reasoning state within the existing context. It estimates the semantic gap between the reasoning state and the accumulated evidence, and strategically alternates between boundary expansion and intra-boundary trajectory refinement. 
This eliminates redundant retrieval steps and preserves a compact, evidence-supported reasoning trajectory.
Extensive experiments on challenging open-domain and multi-hop QA benchmarks demonstrate that EvoCtx significantly outperforms previous methods, offering an effective approach to complex reasoning tasks.
The source code can be accessed at \textcolor{magenta}{\url{https://github.com/Anya-RB-Chen/EvoCtx}.}
\end{abstract}

\section{Introduction}

\begin{figure}[th!]
    \centering
    \includegraphics[width=1\linewidth]{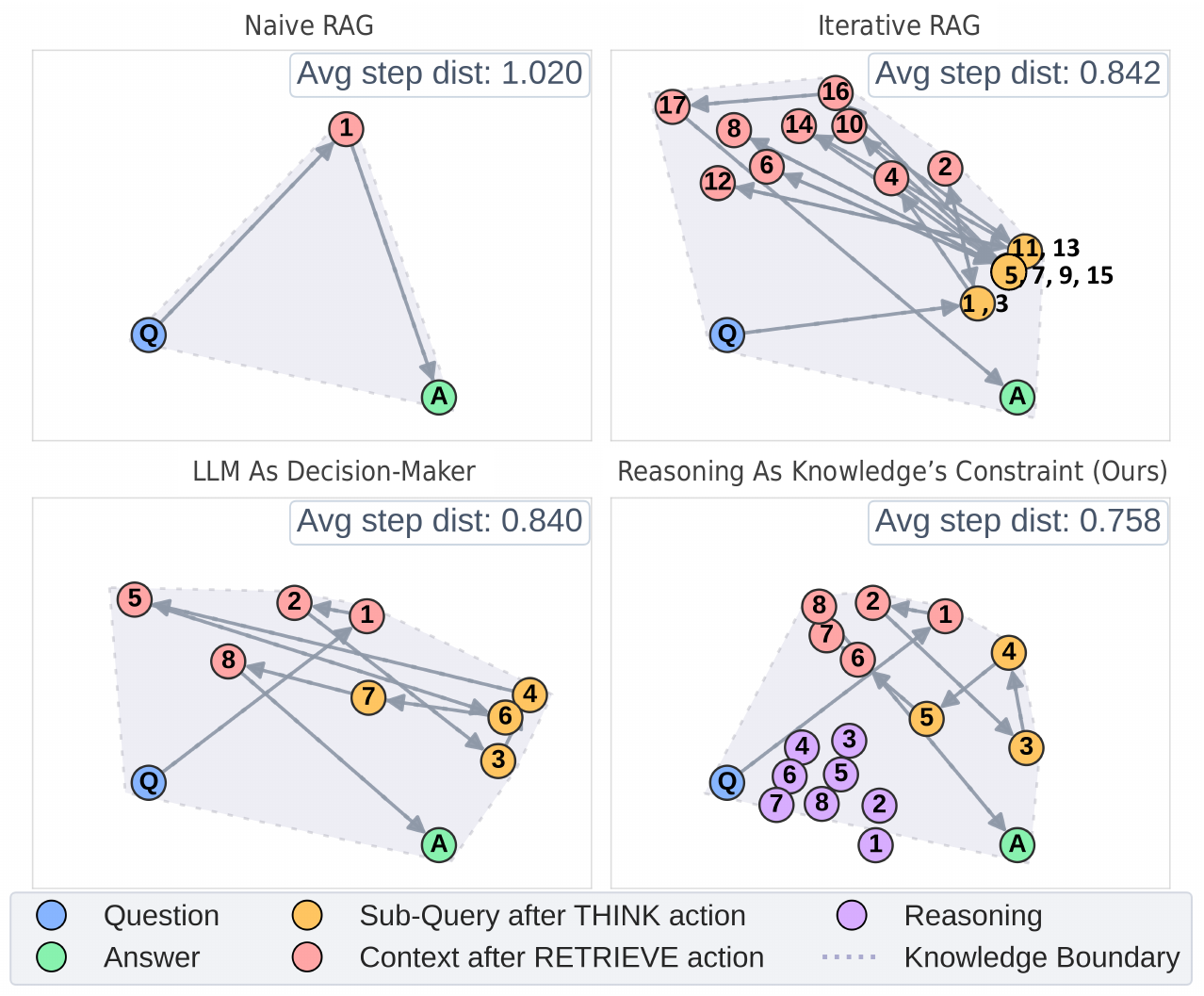}
    \caption{Visualized semantic trajectories of different retrieval-reasoning policies. Examples are selected from the MultiHop-RAG~\cite{tang2024multihop} dataset. Numbered nodes represent intermediate states. Unlike baselines that exhibit coarse jumps (Naive RAG), dispersed states (Iterative RAG), or unsupported steps (LLM as a decision-maker), our \model uses a boundary-aware retrieve-vs.-think controller to decide when to acquire new evidence and when to refine the search trajectory within the current evidence boundary, yielding a compact and evidence-supported trajectory.}
    \label{fig:semantic_trajectory}
    \vspace{-6pt}
\end{figure}

\begin{figure*}[t!]
    \centering
    \includegraphics[width=1\linewidth]{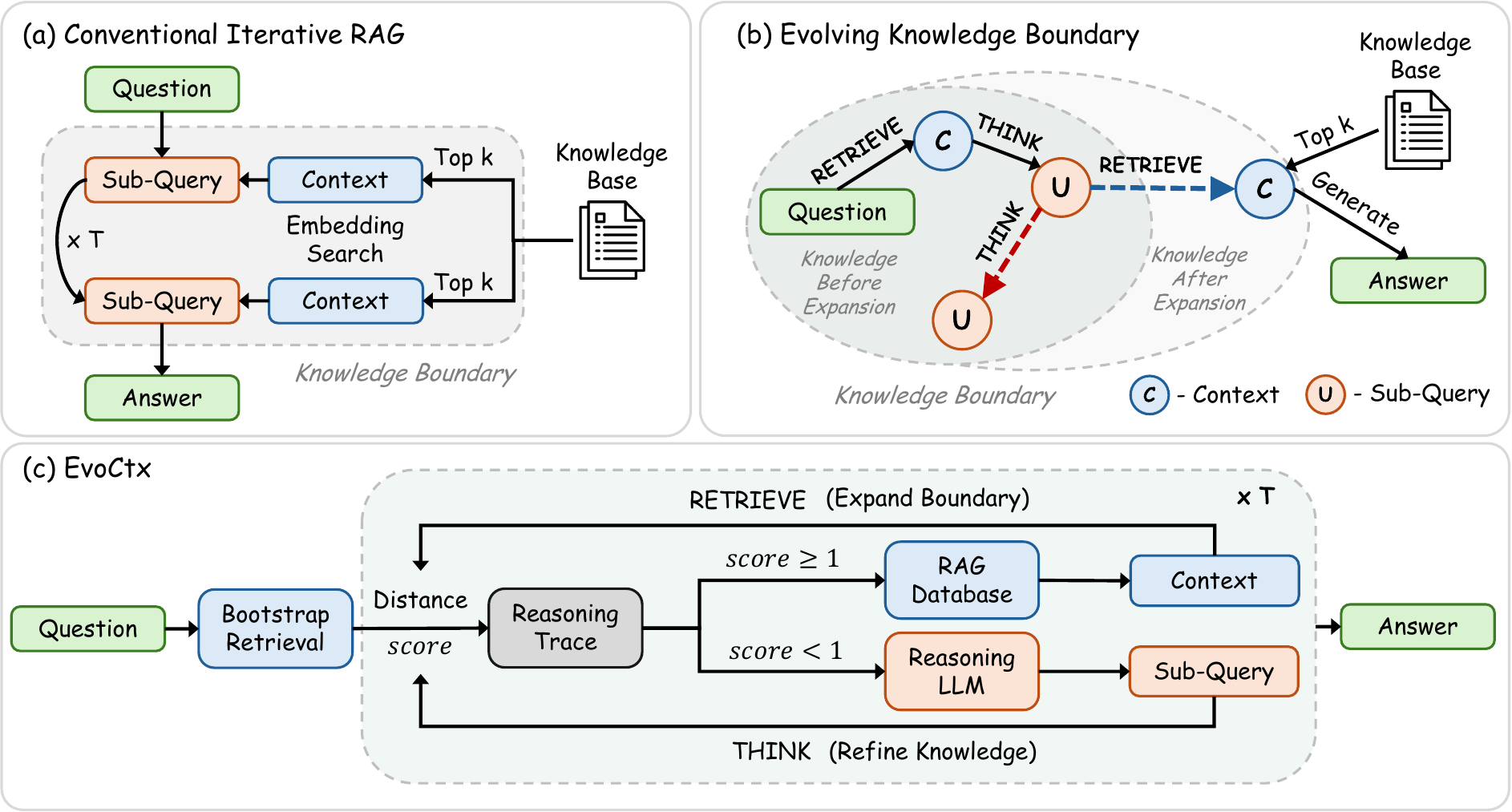}
    \caption{
    Overview of the proposed \textbf{\model} against previous methods. 
    (a) Conventional iterative RAG retrieves at every step, often causing context saturation. 
    (b) Prior knowledge boundary evolving frames reasoning as \emph{dynamic context acquisition}, simply expanding the knowledge boundary step by step. 
    (c) Our \model alternates between boundary expansion and trajectory refinement, yielding a compact, adaptive knowledge state to answer the question.
    }
    \label{fig:framework}
    \vspace{-6pt}
\end{figure*}

Large language models (LLMs) solve complex tasks by conditioning their reasoning and generation on various contexts, which may contain instructions, demonstrations, or retrieved evidence~\cite{brown2020language, wei2022chain}. 
For open-domain complex tasks, retrieval-augmented generation (RAG)~\cite{lewis2020retrieval} is particularly important, as it bridges the gap between internal parametric knowledge and external evidence. 
However, many tasks require multiple hops and evidence is rarely identifiable in a single step. 
Later information needs often become visible only after earlier hops have been resolved~\cite{press2023measuring, trivedi2022interleaving}. 
Therefore, the key challenge is not merely how much context should be provided, but \textit{how context should be acquired, updated, and controlled} as reasoning unfolds~\cite{zhao2024retrieval,wei2012mining, zhang2026mol}.

Conventional context augmentation typically follows a ``retrieve-then-generate'' pipeline, implicitly assuming that the initial query can expose all evidence needed for the final answer~\cite{press2023measuring, trivedi2022interleaving}. 
This assumption is fragile for open-world multi-hop reasoning, where the model must operate under incomplete information and progressively uncover the evidence chain. 
To relax this assumption, recent iterative methods interleave retrieval and generation across multiple rounds~\cite{trivedi2022interleaving, shao2023enhancing, yue2024iterdrag, thompson2025inference, verma2024plan}. 
Despite their improvements, many approaches still treat retrieval as a default action at each step, expanding the context according to a fixed or nearly fixed schedule rather than deciding whether new external evidence is actually necessary.
As shown in Figure~\ref {fig:semantic_trajectory}, a naive RAG pipeline makes a coarse jump from the question to the answer, often leaving intermediate evidence requirements unresolved. Iterative RAG methods~\citep{yue2024iterdrag,li2024kirag} improve coverage by retrieving repeatedly, but their indiscriminate expansion scatters intermediate states with weakly relevant evidence and makes the trajectory unnecessarily long. 
Conversely, allowing an LLM to decide whether retrieval is needed may reduce context size, but can also introduce unsupported transitions when the next hop lies outside the current evidence boundary. 
This suggests that the core problem is not simply to retrieve more or less, but to decide whether each reasoning transition should \emph{cross} the current boundary of accumulated evidence~\cite{wei2012coaching}.

We argue that effective context augmentation for multi-hop reasoning should be framed as \emph{boundary-aware context evolution}. Starting from the input question, the model maintains an accumulated context that defines what information is currently supported by external evidence and prior reasoning. 
When the next reasoning transition requires information beyond this support, the model should retrieve new evidence to expand the boundary. When the transition can be made within the current boundary, the model should instead reason over the existing context, reorganize the accumulated evidence, and refine the unresolved sub-goal~\cite{jiang2026think}. 
This is aligned with human reasoning: effective reasoners do not search indiscriminately, but alternate between seeking new information and making better use of what they already know~\cite{simon1983search, ackerman2017meta}.

Motivated by this perspective, we propose Cross-Boundary \textbf{C}onte\textbf{xt} \textbf{Evo}lution (\textbf{\model}), a framework that turns retrieve-versus-think into a semantic decision over an evolving evidence boundary. 
\model first performs a lightweight bootstrap retrieval to ground the question and establish an initial context. It then iteratively selects between two actions: (\romannumeral1) \texttt{RETRIEVE}, which expands the current boundary with new external evidence; and (\romannumeral2) \texttt{THINK}, which performs intra-boundary reasoning by reorganizing existing evidence, identifying the unresolved semantic gap, and refining the next sub-query without injecting additional contexts. 
Crucially, \model estimates the semantic gap between the reasoning-trace embedding and the knowledge accumulated so far. A large gap indicates that the current context no longer supports the next transition and should be expanded through retrieval, whereas a smaller but unresolved gap suggests that the model should mainly refine its context within the current boundary.

Our main contributions are as follows:
\begin{itemize}
    \item We formulate open-domain multi-hop reasoning as a problem of \emph{boundary-aware context evolution}, shifting the focus from static context construction to adaptive control over how external evidence is gathered.
    \item We propose \textbf{\model}, a simple yet effective framework that applies this formulation through semantic action selection, estimating the gap between the reasoning embedding and accumulated evidence.
    \item Extensive experiments on challenging open-domain and multi-hop QA benchmarks show that \model improves answer accuracy significantly. Further analyses demonstrate that our method yields more compact and evidence-supported reasoning trajectories.
\end{itemize}

\section{Related Work}

\paragraph{Retrieval-Augmented Generation.}
Retrieval-augmented generation (RAG) improves knowledge-intensive generation by grounding language models in external evidence retrieved from a corpus~\cite{lewis2020retrieval,yu2024evaluation}. The original retrieve-then-generate paradigm remains a strong baseline for open-domain and multi-hop QA, but it assumes that the information need is largely exposed by the initial question~\cite{zhu2021retrieving,mavi2024multi}. In multi-hop reasoning, however, missing facts often emerge only after partial reasoning, so a single retrieval step can ground the question but cannot revise the search direction as new evidence needs appear~\cite{lin2024prompting}.

\paragraph{Adaptive Retrieval for Language Models.}
Recent work studies whether retrieval should be invoked selectively rather than unconditionally. Self-RAG uses reflection tokens to let the model retrieve and critique its own generation~\cite{asai2024selfrag}; FLARE retrieves based on predicted uncertain future content~\cite{jiang2023flare}; AdaptiveRAG routes questions to different retrieval strategies by estimated complexity~\cite{jeong2024adaptive}; and DRAGIN triggers retrieval from the model's evolving generation state~\cite{su2024dragin}. These methods share the intuition that retrieval should be controlled rather than blindly appended. 
Different from using surface tokens, question-level routing, or token-level uncertainty, our method decides by comparing the model's reasoning trace with the current knowledge boundary, asking whether the existing boundary still supports progress or must be expanded.

\paragraph{Multi-hop Reasoning.}
Complex QA tasks further highlight the need to interleave retrieval with multi-hop reasoning. IRCoT alternates chain-of-thought reasoning and retrieval so that intermediate reasoning can guide later evidence acquisition~\cite{trivedi2022interleaving}; KiRAG and IterDRAG similarly improve multi-hop RAG through iterative retrieval--reasoning processes~\cite{li2024kirag, yue2024iterdrag}. These methods move beyond one-shot retrieval, but they typically tie retrieval to a procedural loop. We build on the same insight while introducing a stronger decision perspective: open-world RAG should not only evolve context, but decide \emph{how} it should evolve. \model therefore treats inference as dynamic context acquisition and decision-making over an evolving knowledge boundary, choosing between \texttt{RETRIEVE} to expand that boundary and \texttt{THINK} to refine the decision trajectory. 
\section{Methodology}


We propose \textbf{\model}, a framework that formulates retrieval-augmented multi-hop complex reasoning as \emph{dynamic context acquisition and decision making}. Starting from a bootstrap retrieval, \model iteratively chooses between two actions:
\texttt{RETRIEVE}, which expands the current knowledge boundary with new external evidence, and
\texttt{THINK}, which refines the current search trajectory by generating a new sub-query under the existing boundary.
Figure~\ref{fig:framework} shows the overview of \model.

\subsection{Problem Setup}

Given a question \(Q\) and an external corpus \(\mathcal{K}\), \model performs one bootstrap retrieval followed by \(T\) adaptive decision rounds.
Since \(Q\) and \(\mathcal{K}\) remain fixed throughout inference, we define the mutable state at round \(i\) as:
\begin{equation}
\mathcal{S}_i = (\mathcal{C}_i, \mathcal{U}_i),
\end{equation}
where \(\mathcal{C}_i\) is the accumulated retrieved context and \(\mathcal{U}_i\) is the set of generated sub-queries.
The action space is defined as follows:
\begin{equation}
\mathcal{O} = \{\texttt{RETRIEVE}, \texttt{THINK}\}.
\end{equation}
Then, \model begins with a bootstrap retrieval using the original question:
\begin{equation}
\mathcal{C}_0 = f_{\text{ret}}(Q, \mathcal{K}), 
\qquad
\mathcal{U}_0 = \emptyset.
\end{equation}
This step grounds the reasoning process in external evidence before adaptive control begins.

\subsection{Boundary-Aware Decision-Making}

At each round \(i\), a decision model reads the question, current context, and previous sub-queries, and produces a reasoning trace together with an optional surface-form action suggestion:
\begin{equation}
(r_i, \hat{o}_i) = f_{\text{dec}}(Q, \mathcal{C}_i, \mathcal{U}_i).
\end{equation}
Here \(r_i\) is the reasoning trace, while \(\hat{o}_i \in \mathcal{O}\) is the action token produced by the model.
\model does \emph{not} directly execute \(\hat{o}_i\).
Instead, it uses \(r_i\) as a semantic probe of the model's current information need.

\paragraph{Knowledge Boundary.}
To decide whether the current boundary is sufficient, we construct a bounded knowledge boundary as follows:
\begin{equation}
\mathcal{B}_i = \psi(Q, \mathcal{U}_i, \mathcal{C}_i),
\end{equation}
where \(\psi(\cdot)\) collects a deduplicated subset of the evidence accumulated so far, including the original question, previously generated sub-queries, and retrieved passages.
In practice, \(\psi(\cdot)\) can bound the number of sub-queries and passages used for decision making, which keeps the boundary representation compact and prevents old or redundant context dominating the decision signal.

\paragraph{Action Selection with Novelty Score.}

Let \(E(\cdot)\) be a text encoder.
We encode the reasoning trace and the current boundary as:
\begin{equation}
h_i = E(r_i),
\qquad
H_i = \{E(b) \mid b \in \mathcal{B}_i\}.
\end{equation}
\model then measures how far the reasoning embedding moves beyond the current boundary. Let \(\mu_i = |\mathcal{B}_i|^{-1}\sum_{b \in \mathcal{B}_i}E(b)\). We compute three novelty signals:
\begin{align}
d_i^{\text{near}} &= \min_{b \in \mathcal{B}_i} \delta(h_i,E(b)), \\
d_i^{\text{cent}} &= \delta(h_i,\mu_i), \\
d_i^{90} &= \operatorname{Q}_{0.9}\{\delta(h_i,E(b))\}_{b \in \mathcal{B}_i},
\end{align}
where \(\delta(x,y)=1-\cos(x,y)\).

These signals capture different aspects of boundary mismatch.
\(d_i^{\text{near}}\) measures whether the reasoning embedding is unsupported even by its closest boundary item;
\(d_i^{\text{cent}}\) measures whether it drifts away from the global center of the boundary;
and \(d_i^{90}\) captures upper-tail mismatch when the reasoning trace is far from most boundary items.

We aggregate these signals into a threshold-normalized novelty score:
\begin{equation}
 score = \max\!\left\{
\min\!\left(\frac{d_i^{\text{near}}}{\tau_{\text{near}}},
\frac{d_i^{\text{cent}}}{\tau_{\text{cent}}}\right),
\frac{d_i^{90}}{\tau_{90}}
\right\}.
\end{equation}
The executed action is then selected by a compact threshold rule:
\begin{equation}
\label{eq:hybrid_decision}
o_i =
\begin{cases}
\texttt{RETRIEVE}, & score \ge 1, \\
\texttt{THINK}, & score < 1.
\end{cases}
\end{equation}
Therefore, if the reasoning trace lies outside the semantic support of the current boundary, the system should retrieve to expand the boundary;
otherwise, it should think within the current boundary and refine the search trajectory.

\begin{algorithm}[t]
\algrenewcommand\algorithmicrequire{\textbf{Input:}}
\algrenewcommand\algorithmicensure{\textbf{Output:}}
\caption{Inference Process of \model}
\label{alg:ace}
\begin{algorithmic}[1]
\Require Question \(Q\), corpus \(\mathcal{K}\), adaptive round budget \(T\), retriever \(f_{\text{ret}}\), decision model \(f_{\text{dec}}\), sub-query generator \(f_{\text{sub}}\), answer generator \(f_{\text{ans}}\), encoder \(E\), decision rule \(score\).
\Ensure Answer \(f_{\text{ans}} (Q)\).
\State \(\mathcal{U}_0 \gets \emptyset\)
\State \(\mathcal{C}_0 \gets f_{\text{ret}}(Q, \mathcal{K})\) \Comment{bootstrap retrieval}
\For{\(i = 0\) to \(T-1\)}
    \State \((r_i, \hat{o}_i) \gets f_{\text{dec}}(Q, \mathcal{C}_i, \mathcal{U}_i)\)
    \State \(\mathcal{B}_i \gets \psi(Q, \mathcal{U}_i, \mathcal{C}_i)\)
    \State Compute novelty signals \((d_i^{\text{near}}, d_i^{\text{cent}}, d_i^{90})\) from \(E(r_i)\) and \(\{E(b): b \in \mathcal{B}_i\}\)
    \State \(o_i \gets score(d_i^{\text{near}}, d_i^{\text{cent}}, d_i^{90})\)
    \If{\(o_i = \texttt{RETRIEVE}\)}
        \State \(q_i^{\text{ret}} \gets \phi(Q, \mathcal{U}_i)\)
        \State \(D_i \gets f_{\text{ret}}(q_i^{\text{ret}}, \mathcal{K} \setminus \mathcal{C}_i)\)
        \State \(\mathcal{C}_{i+1} \gets \mathcal{C}_i \cup D_i,\quad \mathcal{U}_{i+1} \gets \mathcal{U}_i\)
    \Else
        \State \(u_i \gets f_{\text{sub}}(Q, \mathcal{C}_i, \mathcal{U}_i)\)
        \State \(\mathcal{U}_{i+1} \gets \mathcal{U}_i \cup \{u_i\},\quad \mathcal{C}_{i+1} \gets \mathcal{C}_i\)
    \EndIf
\EndFor
\State \Return \(f_{\text{ans}}(Q, \mathcal{C}_T, \mathcal{U}_T)\)
\end{algorithmic}
\end{algorithm}

\subsection{Context Evolution}

Once the action \(o_i\) is selected, \model updates the context accordingly.

\paragraph{\texttt{RETRIEVE} for Boundary Expansion.}
If \(o_i = \texttt{RETRIEVE}\), \model forms a retrieval query by combining the original question with the accumulated sub-queries, given by:
\begin{equation}
q_i^{\text{ret}} = \phi(Q, \mathcal{U}_i),
\end{equation}
and retrieves new evidence from the corpus as follows:
\begin{equation}
D_i = f_{\text{ret}}(q_i^{\text{ret}}, \mathcal{K} \setminus \mathcal{C}_i),
\end{equation}
where \(D_i\) denotes the newly retrieved passages and previously seen documents are excluded.
The state update is given by:
\begin{equation}
\mathcal{C}_{i+1} = \mathcal{C}_i \cup D_i,
\qquad
\mathcal{U}_{i+1} = \mathcal{U}_i.
\end{equation}
This action explicitly enlarges the semantic coverage of the current knowledge boundary.


\paragraph{THINK for Trajectory Refinement.}
If $o_i=\textsc{THINK}$, a generation model proposes one new sub-query conditioned on the current state:
\begin{equation}
u_i = f_{\text{sub}}(Q, C_i, U_i).
\end{equation}
The state is then updated as:
\begin{equation}
U_{i+1} = U_i \cup \{u_i\}, \quad C_{i+1} = C_i.
\end{equation}
Importantly, \textsc{THINK} does not inject new external evidence or produce an intermediate answer. Its core role is to \textbf{refine the query space under the current evidence boundary}: it reorganizes the next search direction from the currently supported context so that later retrieval, if needed, is better targeted and more answer-relevant. Equivalently, \textsc{RETRIEVE} updates only $C$, while \textsc{THINK} updates only $U$.

\subsection{Final Answer Generation}

After \(T\) adaptive rounds, the final answer is generated from the question, the accumulated context, and the generated sub-queries:
\begin{equation}
A = f_{\text{ans}}(Q, \mathcal{C}_T, \mathcal{U}_T).
\end{equation}
The final prediction is therefore conditioned on a dynamically constructed knowledge state rather than on a single static retrieval result. Algorithm~\ref{alg:ace} summarizes the full \model inference process.

\section{Experiments}
\label{sec:experiments}

\subsection{Experimental Setup}

\paragraph{Benchmarks.}
We evaluate on four QA benchmarks that cover both multi-hop and open-domain evidence-seeking behavior. 2WikiMultihopQA~\cite{xanh2020_2wikimultihop} and HotpotQA~\cite{yang2018hotpotqa} require combining evidence across entities or documents. MultiHop-RAG~\cite{tang2024multihop} further stresses retrieval-augmented multi-hop reasoning in settings where the relevant evidence is distributed across retrieved passages. TriviaQA~\cite{joshi2017triviaqa} provides an additional open-domain QA setting for testing whether a method can acquire useful external evidence beyond multi-hop-only datasets.

\begin{table*}[t!]
\centering
\resizebox{\textwidth}{!}{%
\begin{tabular}{l l | ccc | ccc | ccc | ccc}
\toprule
\multirow{2}{*}{\textbf{Method}} & \multirow{2}{*}{\textbf{Decision Type}} & \multicolumn{3}{c}{\textbf{2Wiki}} & \multicolumn{3}{c}{\textbf{HotPotQA}} & \multicolumn{3}{c}{\textbf{MultiHop-RAG}} & \multicolumn{3}{c}{\textbf{TriviaQA}} \\
 &  & EM & F1 & Acc & EM & F1 & Acc & EM & F1 & Acc & EM & F1 & Acc \\
\midrule
\rowcolor{table_grey} \multicolumn{14}{l}{\textit{Internal Knowledge Only}} \\
Llama-3.1-8B-Instruct & Closed-Book & 0.0 & 5.4 & 48.4 & 0.0 & 4.4 & 49.2 & 7.0 & 7.9 & 37.6 & 0.0 & 8.6 & 69.7 \\
Qwen3-4B-Instruct & Closed-Book & 0.0 & 2.5 & 51.8 & 0.0 & 1.9 & 50.5 & 6.2 & 6.4 & 39.6 & 0.1 & 3.3 & 56.3 \\
\midrule
\rowcolor{table_grey} \multicolumn{14}{l}{\textit{Retrieval-based External Knowledge}} \\
NaiveRAG~\cite{lewis2020retrieval}      & Static   & 30.0 & 38.2 & 48.6 & 38.2 & 51.9 & 56.8 & 36.7 & 39.2 & 45.6 & 53.6 & 62.4 & 60.2 \\
IRCoT~\cite{trivedi2022interleaving}         & Iterative & 36.2 & 46.0 & 45.4 & 41.4 & 51.6 & 43.8 & --   & --   & --   & 52.6 & 63.4 & 60.8 \\
KiRAG~\cite{li2024kirag}         & Iterative & 30.7 & 50.6 & --   & 45.1 & 59.8 & --   & --   & --   & --   & --   & --   & --   \\
IterDRAG~\cite{yue2024iterdrag}      & Iterative & 31.8 & 38.4 & 47.7 & 34.6 & 44.8 & 51.8 & 59.8 & 60.8 & 58.4 & 16.9   & 21.9   & 60.5   \\
AdaptiveRAG~\cite{jeong2024adaptive}   & Adaptive  & 38.4 & 47.1 & 45.4 & 39.6 & 49.9 & 41.4 & --   & --   & --   & 56.4 & 65.6 & 62.8 \\
Self-RAG~\cite{asai2024selfrag}      & Adaptive  & 12.3 & 25.1 & 33.3 & 17.4 & 29.9 & 32.1 & --   & --   & --   & 12.8 & 29.3 & 57.0 \\
FLARE~\cite{jiang2023flare}         & Adaptive  & 35.8 & 45.1 & 42.4 & 29.8 & 39.1 & 37.2 & --   & --   & --   & 57.0 & 67.4 & 64.8 \\
DRAGIN~\cite{su2024dragin}        & Adaptive  & 40.6 & 48.0 & 45.6 & 39.8 & 50.6 & 43.0 & --   & --   & --   & 58.4 & 69.1 & 66.6 \\
\midrule
\textbf{\model (Ours)} & Adaptive + Iterative & \textbf{46.2} & \textbf{50.7} & \textbf{53.6} & \textbf{49.1} & \textbf{63.0} & \textbf{65.4} & \textbf{74.6} & \textbf{74.6} & \textbf{66.3} & \textbf{64.3} & \textbf{72.3} & \textbf{69.8} \\
\bottomrule
\end{tabular}%
}
\caption{Main results comparing \model with baseline methods on complex reasoning benchmarks. We utilize 2Wiki, HotpotQA, and MultiHop-RAG to measure multi-hop reasoning capabilities, while including TriviaQA as a complementary single-hop knowledge-intensive task. Best results are highlighted in \textbf{bold}.
}
\label{tab:main_results}
\vspace{-6pt}
\end{table*}

\paragraph{Implementation Details.}
\model decouples action selection from answer generation. We use Qwen3-4B-Thinking~\cite{qwen3technicalreport} as the decision model that uses its thinking contents' semantics to determine whether the next step should be \texttt{RETRIEVE} or \texttt{THINK}, and LLaMA-3.1-8B-Instruct~\cite{grattafiori2024llama3herdmodels} as the generation model for sub-query generation and final answer prediction. Semantic comparison between the reasoning trace embedding and the current knowledge boundary is implemented with the all-MiniLM-L6-v2 text encoder. The parameter setting has two major variables: the total rounds of the iteration ($T$) and the number of chunks to be retrieved ($k$) is set to be 8 and 5. The boundary thresholds are set as: $\tau_{near}=0.3$, $\tau_{cent}=0.2$, and $\tau_{90}=0.4$.
In the main experiments, we set the adaptive round budget to \(T=8\) and the retrieval depth to top-\(k=5\).

\paragraph{Baseline Methods.}
We compare against the following methods with different decision types: \textbf{Closed-Book} baselines ask LLaMA-3.1-8B-Instruct and Qwen3-4B-Instruct to answer directly from internal knowledge. \textbf{Static} retrieval is represented by NaiveRAG~\cite{lewis2020retrieval}, which performs one retrieval step and then answers from a fixed context. \textbf{Iterative} methods, including IRCoT~\cite{trivedi2022interleaving}, KiRAG~\cite{li2024kirag}, and IterDRAG~\cite{yue2024iterdrag}, repeatedly couple retrieval with intermediate reasoning, but their decision process is largely tied to a predefined loop. \textbf{Adaptive} methods, including AdaptiveRAG~\cite{jeong2024adaptive}, Self-RAG~\cite{asai2024selfrag}, FLARE~\cite{jiang2023flare}, and DRAGIN~\cite{su2024dragin}, decide retrieval more flexibly using complexity, reflection, uncertainty, or information-need signals. 
As an \textbf{Adaptive + Iterative} method, our \model uses reasoning trace embedding to judge whether the current knowledge boundary is sufficient, thereby deciding between \texttt{THINK} and \texttt{RETRIEVE}. The baseline settings are aligned with \model where $T=8$ and $k=5$.

For retrieval-based comparisons, we align shared components whenever the baseline interface is plug-compatible. In particular, we use the same benchmark corpus and final answer generator (LLaMA-3.1-8B-Instruct), and matched upper bounds on the adaptive budget and retrieval depth where these controls are exposed. When a baseline requires its own prompt format, controller, or training recipe by design, we preserve the original implementation rather than forcing artificial standardization. This protocol is intended to keep shared components aligned where meaningful while maintaining the native behavior of each baseline.

\paragraph{Evaluation Metrics.}
Following many existing studies~\cite{lewis2020retrieval,yue2024iterdrag}, we adopt Exact Match (\textbf{EM}), token-level F1 (\textbf{F1}), and Accuracy (\textbf{Acc}) as the evaluation metrics.
EM and F1 measure answer overlap under standard normalization, while Acc measures whether the prediction contains an acceptable ground-truth answer. 

\subsection{Main Results}

Table~\ref{tab:main_results} summarizes the main comparison. Overall, the results support a decision-centric view of RAG: performance improves as methods move from Closed-Book internal knowledge to Static, Iterative, and Adaptive external-knowledge use, but the strongest results require a cognitive decision over whether to \texttt{RETRIEVE} or \texttt{THINK}. \model achieves the best scores across all four benchmarks, with especially large gains on MultiHop-RAG.

\paragraph{External knowledge must be crossed, but not only once.}
Closed-Book baselines obtain near-zero EM on 2Wiki and HotPotQA, showing that internal parametric knowledge is insufficient for exact, evidence-grounded reasoning. NaiveRAG improves substantially by crossing the boundary to external evidence, but the gap between NaiveRAG and \model shows that one static retrieval step cannot handle information needs that emerge after partial reasoning. Iterative methods such as IRCoT~\cite{trivedi2022interleaving}, KiRAG~\cite{li2024kirag}, and IterDRAG~\cite{yue2024iterdrag} address this by repeatedly coupling retrieval with reasoning, yet their retrieval process is still largely procedural. \model improves by making each step conditional on whether the current knowledge boundary semantically supports the reasoning trace embedding.

\paragraph{Boundary-aware control improves over adaptive triggering.}
Adaptive methods use complexity, uncertainty, reflection, or information-need signals to decide retrieval and are strong in several settings, but their performance varies across datasets, suggesting that these signals do not always match the true evidence need. \model advances this adaptive philosophy into a boundary-aware cognitive decision type: it checks whether the current reasoning state is already supported by the accumulated boundary, using \texttt{THINK} to refine the trajectory when supported and \texttt{RETRIEVE} to expand the boundary when unsupported. The largest gain on MultiHop-RAG, where \model reaches 74.6 EM/F1 compared with 59.8/60.8 for IterDRAG, directly supports our central claim that better RAG is not simply more retrieval, but high-level control over when to remain inside the boundary and when to cross it.

\subsection{Ablation Study}

\begin{table}[t!]
\centering
\resizebox{0.7\linewidth}{!}{%
\begin{tabular}{lccc}
\toprule
\textbf{Variant} & EM & F1 & Acc \\
\midrule
\textbf{\model (Ours)} & \textbf{74.6} & \textbf{74.6} & \textbf{66.3} \\
~~~~w/o \texttt{RETRIEVE} & 62.9 & 64.2 & 58.4 \\
~~~~w/o \texttt{THINK} & 71.1 & 71.6 & 64.0 \\
~~~~w/o Reasoning & 63.7 & 64.1 & 58.4 \\
~~~~w/o Boundary & 68.4 & 68.8 & 61.8 \\
\bottomrule
\end{tabular}
}
\caption{Ablation study on the MultiHop-RAG benchmark. Each variant removes one component of \model's dynamic retrieve-or-think decision process.}
\label{tab:ablation}
\end{table}
\begin{table}
\centering
\renewcommand{\arraystretch}{0.85}
\resizebox{0.85\linewidth}{!}{%
\begin{tabular}{c ccccc c}
\toprule
\multirow{2}{*}{\(T\) }& \multicolumn{5}{c}{\textbf{Top-\(k\) Retrieved Passages}} & \multirow{2}{*}{ Avg. } \\
\cmidrule(lr){2-6}
 & 1 & 2 & 3 & 4 & 5 &  \\
\midrule
2 & 60.0 & 55.0 & 60.0 & 60.0 & 64.0 & 59.8 \\
3 & 61.0 & 57.0 & 66.0 & 64.0 & 67.0 & 63.0 \\
4 & 59.0 & 58.0 & 59.0 & 68.0 & 63.0 & 61.4 \\
5 & 56.0 & 59.0 & 58.0 & 63.0 & 65.0 & 60.2 \\
6 & 55.0 & 65.0 & 64.0 & 63.0 & 68.0 & 63.0 \\
7 & 60.0 & 56.0 & 60.0 & 62.0 & 71.0 & 61.8 \\
8 & 60.0 & 65.0 & 66.0 & 68.0 & \textbf{72.0} & \textbf{66.2} \\
\midrule
Avg. & 58.7 & 59.3 & 61.9 & 64.0 & \textbf{67.1} & -- \\
\bottomrule
\end{tabular}%
}
\caption{Sensitivity of \model to the adaptive round budget \(T\) and the number of retrieved passages \(k\) on a 100-example MultiHop-RAG subset. Each value denotes EM (\%). The best setting is highlighted in bold.}
\label{tab:ablation_budget}
\vspace{-6pt}
\end{table}

\begin{table}[t]
\centering
\small
\setlength{\tabcolsep}{3pt}
\resizebox{\linewidth}{!}{%
\begin{tabular}{lccc c@{\quad}lccc c}
\toprule
\multicolumn{5}{c}{\textbf{Hybrid / Tail-Based Rules}} & \multicolumn{5}{c}{\textbf{Single-Signal Rules}} \\
\cmidrule(lr){1-5}\cmidrule(lr){6-10}
\textbf{Rule} & $\tau_{near}$ & $\tau_{cent}$ & $\tau_{90}$ & \textbf{EM} & \textbf{Rule} & $\tau_{near}$ & $\tau_{cent}$ & $\tau_{90}$ & \textbf{EM} \\
\midrule
hybrid       & 0.3 & 0.2 & 0.4 & \textbf{72.0} & nearest  & 0.3 & --  & --  & 63.0 \\
hybrid       & 0.4 & 0.3 & 0.5 & 64.0           & nearest  & 0.4 & --  & --  & 66.0 \\
hybrid       & 0.5 & 0.4 & 0.6 & 68.0           & nearest  & 0.5 & --  & --  & 62.0 \\
\addlinespace[2pt]
distribution & --  & --  & 0.4 & 68.0           & centroid & --  & 0.2 & --  & 67.0 \\
distribution & --  & --  & 0.5 & 64.0           & centroid & --  & 0.3 & --  & 58.0 \\
distribution & --  & --  & 0.6 & 54.0           & centroid & --  & 0.4 & --  & 62.0 \\
\bottomrule
\end{tabular}%
}
\caption{Threshold sensitivity on the MultiHop-RAG test subset (100 examples, $N$=8, $K$=5). ``\texttt{--}'' denotes that the entries are unused.}
\label{tab:ablation_threshold}
\vspace{-6pt}
\end{table}

Table~\ref{tab:ablation} shows that each part of \model contributes to the final performance. 
\textbf{w/o \texttt{RETRIEVE}} disables evidence acquisition after the bootstrap step, so the model can only continue refining its trajectory using \texttt{THINK}. Its large drop confirms that bootstrap evidence alone cannot satisfy later-emerging information needs in open-world multi-hop reasoning. 
\textbf{w/o \texttt{THINK}} removes intermediate trajectory refinement and forces every subsequent step to retrieve new evidence, namely to act \texttt{RETRIEVE}. The resulting degradation shows that blind boundary expansion is less effective than selectively reasoning within the current boundary before retrieving again. 
\textbf{w/o Reasoning} replaces the reasoning-based decider with LLaMA-3.1-8B and treats the control decision as a shallow action prediction without explicit reasoning trace. Its weaker performance indicates that the model must expose its current information need through reasoning in order to decide whether the boundary is sufficient. 
\textbf{w/o Boundary} still uses the reasoning LLM, but directly executes its generated action token without checking whether the reasoning embedding is semantically supported by the accumulated boundary. The drop relative to full \model demonstrates that semantic verification is necessary to turn reasoning into a reliable boundary-control signal.

\paragraph{Sensitivity to the round budget \(T\) and retrieval depth \(k\).}
Table~\ref{tab:ablation_budget} examines how \model behaves under different adaptive round budgets and retrieval depths. Two patterns are clear. First, larger \(k\) is consistently beneficial: the average EM rises from 58.7 at \(k=1\) to 67.1 at \(k=5\), indicating that when the controller decides to cross the current knowledge boundary, expanding it with sufficiently rich evidence is important. Second, increasing \(T\) is most helpful when paired with adequate retrieval capacity, with the best result achieved at \(T=8\) and \(k=5\). At the same time, the gains are not strictly monotonic for every setting, which supports our central claim: performance does not come from blindly adding more iterations or more retrieved passages, but from enabling effective cognitive control over \texttt{THINK}-versus-\texttt{RETRIEVE} decisions as the knowledge boundary evolves.

\paragraph{Threshold sensitivity.}
Table~\ref{tab:ablation_threshold} examines how different threshold rules instantiate \model's semantic boundary on a 100-example MultiHop-RAG subset. The hybrid rule performs best, reaching 72 EM under permissive thresholds, suggesting that \model benefits from retrieving as soon as multiple novelty signals jointly indicate that the current frontier is insufficient. By contrast, single-signal rules are weaker and less stable: nearest and centroid never match the hybrid peak, while the distribution-only rule degrades as \(\tau_{90}\) becomes more conservative. Overall, this supports our design: cognitive control is more reliable when novelty is judged from complementary geometric signals.

\subsection{In-depth Analysis}

\begin{figure}[t!]
    \centering
    \includegraphics[width=1\linewidth]{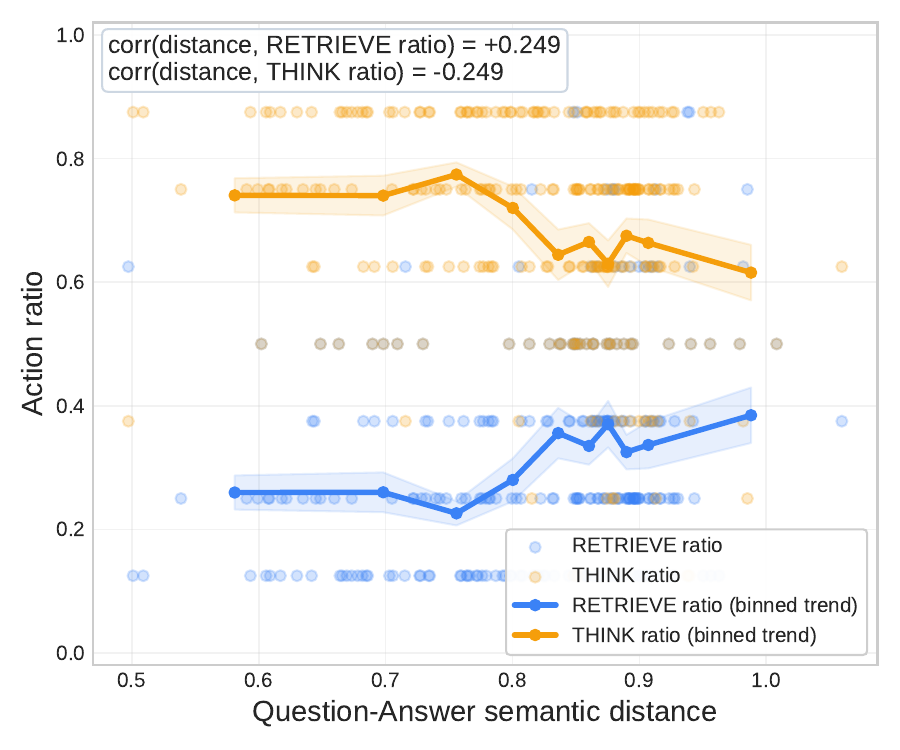}
    \caption{Action allocation with respect to question--answer semantic distance. Each point reports the ratio of \texttt{RETRIEVE} or \texttt{THINK} actions for one example, while solid curves show binned trends. Larger semantic distance is associated with a higher retrieval ratio and a lower thinking ratio, suggesting that \model adaptively expands the knowledge boundary more often when the answer is semantically farther from the question.}
    \label{fig:action_ratio_distance}
    \vspace{-6pt}
\end{figure}

\paragraph{Action allocation follows semantic difficulty.}
Figure~\ref{fig:action_ratio_distance} analyzes whether \model's decisions reflect example-level semantic difficulty rather than following a fixed retrieval schedule. We use the question--answer semantic distance as a proxy for how far the reasoning process must move from the initial question to the target ground truth. The positive correlation between distance and \texttt{RETRIEVE} ratio, together with the symmetric negative correlation for \texttt{THINK}, shows that \model tends to retrieve more when the answer lies farther beyond the initial semantic neighborhood. This supports our boundary-crossing interpretation, where harder cases require more frequent expansion of the knowledge boundary. Importantly, the trend is moderate rather than absolute; even high-distance examples retain substantial \texttt{THINK} actions. This indicates that \model does not reduce difficulty to simply retrieving more documents. Instead, it coordinates retrieval with trajectory refinement: \texttt{RETRIEVE} expands the boundary when evidence is missing, while \texttt{THINK} keeps the search direction coherent so that later retrieval is better targeted.

\begin{figure}[t!]
    \centering
    \includegraphics[width=1\linewidth]{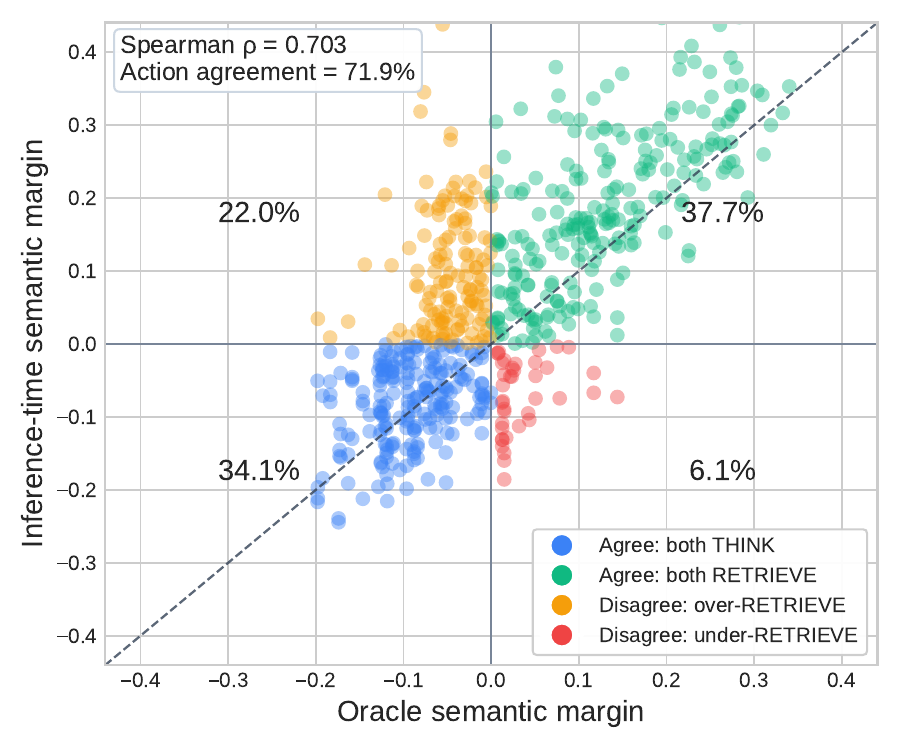}
    \caption{Inference-time versus post hoc oracle semantic margin at each \model decision round. Each point is one round under the same knowledge boundary. Positive margins indicate \texttt{RETRIEVE} and negative margins indicate \texttt{THINK}. The strong correlation and 71.9\% action agreement show that \model's reasoning signal usually matches the oracle decision on whether to expand the boundary or continue reasoning within it.}
    \label{fig:margin_agreement}
    \vspace{-6pt}
\end{figure}

\begin{table*}[t]
\centering
\small
\begin{tabular}{lccc}
\toprule
Method & Avg. retrieval calls & Avg. retrieved passages & Avg. tokens consumed \\
\midrule
NaiveRAG & 1.0 & 5.0 & 1.1k \\
IterDRAG & 8.0 & 17.6 & 18.2k \\
EvoCtx (Ours) & 3.2 & 12.8 & 27.5k \\
\bottomrule
\end{tabular}
\caption{Inference-time efficiency on MultiHop-RAG. EvoCtx uses fewer retrieval calls and fewer retrieved passages than fixed iterative retrieval, while consuming more total tokens due to reasoning traces and \textsc{THINK} steps.}
\label{tab:efficiency}
\end{table*}

\paragraph{Inference-time decisions align with oracle evidence needs.}
Figure~\ref{fig:margin_agreement} evaluates whether \model's inference-time decision signal is a faithful proxy for the true need for external evidence. For each decision round, we compare the signed semantic margin ($score-1$) computed from the model's reasoning trace with a post hoc oracle semantic margin computed from ground-truth target semantics under the same knowledge boundary. The strong positive correlation (\(\rho = 0.703\)) shows that when the reasoning state moves farther beyond the current boundary, the oracle assessment of missing information also tends to increase. Moreover, \(71.9\%\) of decision points fall in the agreement quadrants, indicating that \model usually makes the same \texttt{RETRIEVE}-versus-\texttt{THINK} choice as the oracle. The disagreement pattern is also asymmetric: the more concerning under-\texttt{RETRIEVE} region, where the oracle indicates that the boundary should be expanded but \model remains in \texttt{THINK}, accounts for only \(6.1\%\) of cases, while most disagreement comes from over-\texttt{RETRIEVE} decisions. Despite over-\texttt{RETRIEVE} being able to somehow incorporate noisy data into the knowledge, the robustness for the question-answering task is stable due to enough information gain. This pattern is consistent with our main claim that effective open-world RAG is not simply repeated retrieval, but cognitive control over when to cross the current knowledge boundary and when to continue refining the reasoning trajectory.

\paragraph{Efficiency and practical tradeoffs.}
EvoCtx is designed to improve retrieval allocation rather than to minimize latency at all costs. Table~\ref{tab:efficiency} shows that, on MultiHop-RAG, EvoCtx uses substantially fewer retrieval calls than fixed iterative retrieval and retrieves fewer total passages, indicating a more selective expansion of the evidence boundary. At the same time, EvoCtx can consume more total tokens because \textsc{THINK} steps expose reasoning traces and refine sub-queries before later retrieval. This tradeoff is consistent with our goal: EvoCtx is retrieval-efficient relative to fixed iterative retrieval, while remaining more computation-heavy than one-shot static RAG.
\section{Conclusion}

We presented \model, a framework that formulates retrieval-augmented multi-hop reasoning as \emph{dynamic context acquisition and decision making} over an evolving knowledge boundary. Instead of treating retrieval as the default action, \model decides whether to \texttt{RETRIEVE} or \texttt{THINK} by measuring the semantic gap between the model's reasoning trace and the knowledge accumulated so far. 
%
Experiments on challenging multi-hop and open-domain QA benchmarks show that this boundary-aware retrieve-versus-think control improves performance over static, iterative, and adaptive baselines.
Our results support a claim that better open-world RAG is not only about retrieving more evidence, but about deciding when the current boundary should be expanded and when reasoning should continue within it.

\section*{Limitations}


Our evaluation is conducted on publicly available question answering benchmarks and under a standard text-retrieval setting, which together provide a relatively clean, controlled, and reproducible testbed for studying metacognitive retrieve-versus-think decisions. This setup allows us to examine the proposed mechanism without introducing additional confounding factors from more specialized environments or heterogeneous retrieval conditions. While the selected datasets already cover both multi-hop reasoning and open-domain evidence-seeking behavior, they still represent only a limited portion of the broader landscape of knowledge-intensive tasks. Therefore, further validation on a wider range of domains would help strengthen the evidence for the framework’s robustness and improve confidence in its generality across more diverse application scenarios.

\section*{Ethics Statement}

This work is intended to improve the reliability of retrieval-augmented reasoning. Our experiments use public benchmarks and standard open-sourced language models, and we do not collect human subject data. Like other retrieval-augmented systems, the method may inherit inaccuracies, omissions, or social biases from the underlying language models and external corpora. However, by explicitly deciding when new evidence is needed and by maintaining an evidence-supported reasoning trajectory, \model can provide a more transparent basis for knowledge acquisition than fixed retrieval loops. For real-world deployment, standard safeguards such as corpus curation, source tracking, and human oversight remain important, especially in sensitive application settings.

\section*{Acknowledgements}
This work was supported by the National Natural Science Foundation of China (NSFC) under Grant No. T2541073 and No. 62372314. The experimental part of this work was supported by The Centre for Large AI Models (CLAIM) of The Hong Kong Polytechnic University.

\bibliography{reference}

\newpage
\appendix
\section{Additional Robustness Analyses}
\label{app:robustness}

This appendix reports additional robustness results that are omitted from the main paper due to space constraints. These experiments are designed to test whether EvoCtx's gain depends heavily on a particular threshold choice, signal combination, or representation backend. Overall, the appendix supports the same conclusion as the main paper: EvoCtx is not merely benefiting from one lucky configuration, but from a stable boundary-aware decision mechanism.

\subsection{Cross-dataset Threshold Robustness}

To examine whether the controller is overly sensitive to the exact threshold configuration, we compare several rule families on two datasets and report the best setting within each family. Table~\ref{tab:threshold_summary} is intended to answer two questions: (1) whether the full hybrid controller is consistently strong across datasets, and (2) whether simplified rules can recover most of the gain. The results show that the full hybrid rule remains the strongest on both MultiHop-RAG and HotPotQA, while simpler variants remain competitive but consistently weaker. 
This supports our claim that the three signals provide complementary evidence for boundary-aware control, rather than the performance being driven by a single threshold heuristic.

\begin{table}[ht]
\centering
\small
\begin{tabular}{llc}
\toprule
Dataset & Rule family / best setting & Best EM \\
\midrule
MultiHop-RAG & Hybrid (0.3, 0.2, 0.4) & 72.0 \\
MultiHop-RAG & Near + Tail (0.5, 0.4) & 71.0 \\
MultiHop-RAG & Near + Centroid (0.3, 0.2) & 70.0 \\
MultiHop-RAG & Centroid + Tail (0.2, 0.5) & 66.0 \\
MultiHop-RAG & Nearest-only (0.4) & 66.0 \\
MultiHop-RAG & Centroid-only (0.2) & 67.0 \\
MultiHop-RAG & Tail-only (0.4) & 68.0 \\
\midrule
HotPotQA & Hybrid (0.3, 0.3, 0.4) & 49.0 \\
HotPotQA & Near + Tail (0.3, 0.4) & 48.0 \\
HotPotQA & Near + Centroid (0.4, 0.3) & 44.0 \\
HotPotQA & Centroid + Tail (0.3, 0.5) & 46.0 \\
HotPotQA & Nearest-only (0.5) & 45.0 \\
HotPotQA & Centroid-only (0.3) & 39.0 \\
HotPotQA & Tail-only (0.4) & 41.0 \\
\bottomrule
\end{tabular}
\caption{Best threshold settings across datasets and rule families. The full hybrid rule remains strongest, while simplified variants remain competitive.}
\label{tab:threshold_summary}
\end{table}

\subsection{Pairwise Signal Ablations}

We further isolate the contribution of each signal by retaining only pairwise combinations of the three novelty signals. Table~\ref{tab:pairwise_signals} complements Table~\ref{tab:threshold_summary} by testing whether all three signals are necessary, or whether two are already sufficient. The strongest pairwise controller is \emph{Nearest + Tail}, suggesting that local support and broad mismatch together capture most of the useful decision signal. However, the full hybrid controller still performs best overall, which supports our design choice of using all three signals for the most reliable boundary judgment.

\begin{table}[h]
\centering
\small
\begin{tabular}{lcc}
\toprule
Signals kept & Best setting & EM \\
\midrule
Nearest + Centroid & (0.3, 0.2) & 70.0 \\
Nearest + Tail & (0.5, 0.4) & 71.0 \\
Centroid + Tail & (0.2, 0.5) & 66.0 \\
Full Hybrid & (0.3, 0.2, 0.4) & 72.0 \\
\bottomrule
\end{tabular}
\caption{Pairwise signal ablations on MultiHop-RAG. The nearest-plus-tail variant is the strongest pairwise controller, but the full three-signal rule remains best.}
\label{tab:pairwise_signals}
\end{table}

\subsection{Reasoning-Trace Model and Boundary Encoder Attribution}

The main paper uses \textsc{Qwen3-4B-Thinking} to produce the reasoning trace and \textsc{all-MiniLM-L6-v2} to encode the current boundary. To test whether EvoCtx is overly tied to this exact combination, Table~\ref{tab:model_attribution} varies both the reasoning-trace model and the boundary encoder while keeping the final answer generator fixed. The table is intended to separate two sources of gain: the quality of the reasoning trace and the quality of the boundary representation. The best performance is achieved by the original configuration, indicating that explicit reasoning traces and a lightweight sentence encoder are a good match for our controller. At the same time, the overall trend remains stable across alternative combinations, suggesting that the benefit of EvoCtx comes from the boundary-aware control mechanism itself rather than from a single fragile representation choice.

\begin{table}[h]
\centering
\small
\begin{tabular}{ccc}
\toprule
Adaptive round \(T\) & Avg. confidence score & Delta \\
\midrule
2 & 0.49 & -- \\
3 & 0.56 & +0.07 \\
4 & 0.72 & +0.16 \\
5 & 0.70 & -0.02 \\
6 & 0.78 & +0.08 \\
7 & 0.81 & +0.03 \\
8 & 0.82 & +0.01 \\
\bottomrule
\end{tabular}
\caption{Confidence probing on HotPotQA. The score is the logit probability of ``Yes'' when the model is asked whether the current evidence is sufficient for answering the question. Overall, the evolving context becomes increasingly answerable across steps.}
\label{tab:confidence_probing}
\end{table}

\begin{table*}[t]
\centering
\small
\begin{tabular}{l l c c c}
\toprule
Reasoning trace model & Boundary encoder & EM & F1 & Acc \\
\midrule
Qwen3-4B-Thinking & all-MiniLM-L6-v2 & 75.0 & 75.0 & 66.0 \\
Qwen3-4B-Instruct & all-MiniLM-L6-v2 & 68.0 & 61.0 & 65.0 \\
LLaMA-3.1-8B-Instruct & all-MiniLM-L6-v2 & 66.0 & 70.0 & 69.0 \\
Qwen3-4B-Thinking & LLaMA hidden states & 71.0 & 70.0 & 67.0 \\
Qwen3-4B-Instruct & LLaMA hidden states & 63.0 & 65.0 & 61.0 \\
LLaMA-3.1-8B-Instruct & LLaMA hidden states & 60.0 & 74.0 & 68.0 \\
\bottomrule
\end{tabular}
\caption{Model attribution on MultiHop-RAG with the final answer generator fixed to LLaMA-3.1-8B-Instruct. The overall trend is preserved across reasoning-trace models and boundary encoders.}
\label{tab:model_attribution}
\end{table*}

\section{Additional Experimental Details}
\label{app:details}

This section provides implementation details that are summarized only briefly in the main paper.

\subsection{Full Implementation Details}

Table~\ref{tab:impl_details} lists the main implementation choices of EvoCtx, including retrieval backend, chunking strategy, indexing, prompt roles, and normalization. 
It specifies the exact inference-time setup under which the reported numbers are obtained and highlights that EvoCtx is a training-free controller built on top of standard retriever and generator components.

\begin{table*}[t]
\centering
\small
\begin{tabular}{p{0.20\textwidth} p{0.72\textwidth}}
\toprule
Component & Setting \\
\midrule
Retriever & Dense retriever based on \textsc{all-MiniLM-L6-v2}, using cosine / inner-product similarity over a FAISS flat index. \\
Corpus and chunking & For 2WikiMultiHopQA and HotPotQA, we follow the benchmark-provided corpus processing. For MultiHop-RAG and TriviaQA, we use sentence-aware chunking with \texttt{SentenceSplitter}, 256-token chunks, and 32-token overlap. \\
Indexing & Passage embeddings are precomputed once, normalized, and indexed with flat indexing. \\
Retrieval depth & Bootstrap retrieval and each \textsc{RETRIEVE} step use top-\(k=5\) passages in the main experiments. \\
Top-\(k\) handling & Duplicate passages are removed; previously seen documents are excluded; retrieved passages are appended in retrieval-score order and truncated to a 1024-token context budget. \\
Decision prompt & The prompt asks the decision model to expose its reasoning trace over the current question, accumulated context, and generated sub-queries. \\
Sub-query prompt & The prompt asks the generator to produce one unresolved next-hop sub-query under the current evidence boundary. \\
Final answer prompt & The prompt asks the answer model to produce a short-form answer from the final question-context-query state. \\
Answer normalization & Standard EM/F1 normalization: lowercasing, punctuation/article removal, whitespace normalization, plus dataset-specific aliases where applicable. \\
Training & None. EvoCtx is inference-time only. \\
\bottomrule
\end{tabular}
\caption{Full implementation details of EvoCtx.}
\label{tab:impl_details}
\end{table*}

\subsection{Baseline Alignment Policy}

A common concern in RAG comparisons is whether the observed gain comes from the controller itself or from differences in retrievers, prompts, and answer backbones. Table~\ref{tab:baseline_alignment} clarifies our alignment policy for baseline comparisons. The table is intended to document which components are standardized whenever possible and which are left unchanged to preserve the native behavior of the compared method. This protocol supports the claim that our comparison is controlled where meaningful, while avoiding artificial modifications that would disadvantage baseline methods.

\begin{table*}[t]
\centering
\small
\begin{tabular}{p{0.22\textwidth} p{0.68\textwidth}}
\toprule
Aspect & Treatment \\
\midrule
Corpus & The same benchmark corpus is used for all methods. \\
Final answer model & We use the same final answer generator (\textsc{LLaMA-3.1-8B-Instruct}) for retrieval-based comparisons whenever the baseline interface is plug-compatible; exceptions are preserved only when required by the original method design. \\
Retriever / index & The same retriever backend is used when the baseline is plug-compatible; otherwise, we preserve the original implementation. \\
Adaptive budget / retrieval depth / context budget & We match the same upper budget whenever the baseline exposes these knobs. \\
Prompt style & Baseline-specific when required by the original method design. \\
Training & We preserve the original baseline training setup; EvoCtx itself is training-free. \\
Controller parameters & EvoCtx's controller thresholds are method-specific and are selected on validation data before test-time evaluation. \\
\bottomrule
\end{tabular}
\caption{Comparison protocol used to align shared components while preserving the native behavior of each baseline.}
\label{tab:baseline_alignment}
\end{table*}

\section{Additional Evidence for the Decision Signal}
\label{app:signal}

Beyond end-task accuracy, we also provide supporting evidence that EvoCtx's controller tracks whether the evolving context is becoming more answer-ready over time. 

\subsection{Confidence Probing Across Reasoning Steps}

To measure whether the accumulated context becomes increasingly sufficient for answering the question, we probe the model at different adaptive rounds with a binary sufficiency question and record the confidence assigned to the positive response. Table~\ref{tab:confidence_probing} shows that the average confidence generally rises as more retrieve-or-think steps are performed. This supports the claim that EvoCtx is not only collecting more information, but is progressively organizing the context into a more answerable knowledge state. The small local fluctuation is expected because some intermediate steps prioritize sub-query refinement over immediate answerability, but the overall trend remains upward.

\end{document}